\documentclass{article}
\usepackage{spconf,amsmath,graphicx}
\usepackage{multirow}
\usepackage{comment}
\usepackage{color}
\usepackage{makecell}

\usepackage{lipsum}
\DeclareMathOperator*{\argmax}{arg\,max}

\let\OLDthebibliography\thebibliography
\renewcommand\thebibliography[1]{
  \OLDthebibliography{#1}
  \setlength{\parskip}{0pt}
}

\title{Improved Audio Embeddings by Adjacency-based Clustering with Applications in Spoken Term Detection}
\name{Sung-Feng Huang, Yi-Chen Chen, Hung-yi Lee, Lin-shan Lee}
\address{National Taiwan University, Taiwan}

\begin{document}

\ninept
\maketitle
%
\begin{abstract}
Embedding audio signal segments into vectors with fixed dimensionality is attractive because all following processing will be easier and more efficient, for example modeling, classifying or indexing.
Audio Word2Vec previously proposed was shown to be able to represent audio segments for spoken words as such vectors carrying information about the phonetic structures of the signal segments.
However, each linguistic unit (word, syllable, phoneme in text form) corresponds to unlimited number of audio segments with vector representations inevitably spread over the embedding space, which causes some confusion.
It is therefore desired to better cluster the audio embeddings such that those corresponding to the same linguistic unit can be more compactly distributed.
In this paper, inspired by Siamese networks, we propose some approaches to achieve the above goal.
This includes identifying positive and negative pairs from unlabeled data for Siamese style training, disentangling acoustic factors such as speaker characteristics from the audio embedding, handling unbalanced data distribution, and having the embedding processes learn from the adjacency relationships among data points. 
All these can be done in an unsupervised way.
Improved performance was obtained in preliminary experiments on the LibriSpeech data set, including clustering characteristics analysis and applications of spoken term detection.
\end{abstract}

%

\section{Introduction}
\label{sec:intro}
Speech recognition technologies have been very successful today. 
But for good accuracy in real-world applications, machines still have to learn from huge quantities of annotated data.
This makes the development of speech technologies for a new language challenging.
For low-resourced languages, collecting huge quantities of data is difficult, and having them annotated is even prohibitively hard.
But more then 95\% of the languages all over the world are low-resourced, and many of them even without linguistic analysis, or without written forms.
Compared to annotating audio data, obtaining unannotated audio data of reasonable size is relatively achievable.
If the machines can automatically learn the acoustic patterns for linguistic units (words, syllables, phonemes, etc.) within the speech signals from an unannotated speech data set of reasonable size, recognition models for those units may be constructed, and speech recognition may become possible for a new language under a new environment with minimum supervision.
Imagine a Hokkien-speaking family obtaining an intelligent device at home: this device does not know Hokkien at all in the beginning, by hearing people speaking Hokkien for some time, it may automatically learn the language. 
The goal of this paper is one step forward towards this vision~\cite{CLSP12}.

With the above long-term goal in mind, it is highly desired to embed audio signal segments (probably correspond to linguistic units of words, syllables or phonemes) of variable length into vectors of fixed dimensionality, serving as the latent representations for the signal segments~\cite{he2016multi,settle2016discriminative,chung2016audio,bengio2014word,levin2013fixed,settle2017query,cho2014learning,hsu2017unsupervised,jansen2017unsupervised,karita2018semi}.
This naturally makes all following processing easier such as computation, clustering, modeling classification, indexing, etc.
Good application examples include speaker identification~\cite{dehak2009support}, emotion classification~\cite{schuller2009interspeech}, and spoken term detection (STD)~\cite{lee2013enhanced,chen2013hybrid,norouzian2012exploiting,ram2018cnn,yuan2018learning,shankarspoken}.
In these applications, standard processing mechanisms can be easily performed over such vector representations for the audio segments, achieving the purpose much more efficiently than processing over the signal segments of variable lengths~\cite{bengio2014word,levin2013fixed,kamper2016deep,levin2015segmental,chen2015query}.

Audio Word2Vec was proposed and can be trained in an unsupervised way with a sequence-to-sequence autoencoder, with the embeddings for the input audio segments extracted from the bottleneck layer~\cite{chung2016audio,chen2013hybrid}.
It has been shown that the vector representations obtained in this way carry the phonetic information about the audio segments~\cite{chen2013hybrid}.
It was then further shown that dividing utterances into audio segments and embedding them as sequences of vectors can be jointly learned in an unsupervised way in Segmental Audio Word2Vec~\cite{wang2018segmental}.
Such unsupervised approaches for audio segment embedding are attractive because no annotation is needed.
However, each linguistic unit (word, syllable, phoneme) corresponds to unlimited number of audio realizations each with its own vector representations, and the spread of these vector representations inevitably lead to confusion, especially when no human labels are available.
For example, although embeddings for the realizations of the word "brother" are very close to each other in the vector space, so do those for the word "bother", spread of those two groups causes some confusion.
A Siamese convolutional neural network~\cite{hadsell2006dimensionality,mueller2016siamese,yih2011learning,shaham2018learning} was trained using side information to obtain embeddings for which same-word pairs were closer and different-word pairs were better separated. 
But human annotation is required under this supervised learning scenario~\cite{kamper2016deep}.

Siamese networks learning from same-word and different-word pairs~\cite{kamper2016deep} can be useful in learning better audio embeddings for linguistic units which are discrete, but labeled data is needed. %
In this paper, inspired by the concept of Siamese networks, we propose a set of approaches to learn better audio embeddings based on the adjacency relationships among data points.
This includes identifying positive and negative pairs from unlabeled data for Siamese style training, disentangling acoustic factors such as speaker characteristics from the audio embedding, and handling the unbalanced data distribution.
All these can be done in an unsupervised way, and very encouraging results were observed in the initial experiments.

\section{Proposed Approach} 
\label{sec:method}

Because the goal of improving audio embedding is challenging, in the initial effort here we slightly simplify the task by assuming all training utterances have been properly segmented into spoken linguistic units (words, syllables, phonemes).
Many approaches for segmenting utterances automatically have been developed~\cite{wang2018segmental}, and automatic segmentation plus audio embedding has been jointly trained successfully and reported before~\cite{wang2018segmental}, so such an assumption is reasonable here.

Below we denote the audio corpus as $\mathbf{X} = {\{\mathbf{x}_{i}\}}_{i=1}^{M}$, which consists of $M$ spoken linguistic units, each represented as an acoustic feature sequence of length $T$, $\mathbf{x}_i=(\mathbf{x}_{i_1}, \mathbf{x}_{i_2}, ..., \mathbf{x}_{i_T})$.
In the subsections below, we try to perform the Siamese style training in an unsupervised way over audio segments.

\begin{figure}[t]
  \centering
  \includegraphics[width=0.8\linewidth]{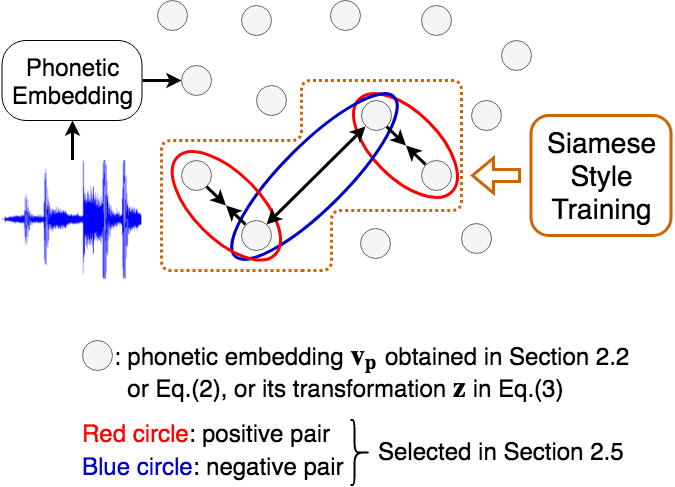}
  \caption{Overview of the proposed approach.}
  \label{fig:overview}
\end{figure}

\subsection{Siamese Networks Considered for Unlabeled Audio Data} 
\label{subsec:stage2}
The overview of the proposed approach is shown in Fig.~\ref{fig:overview}.
Siamese networks are typically trained on a collection of positive and negative pairs of data points to make sure positive pairs are closer and negative pairs far apart.
We wish to use such a concept to improve the audio embeddings considered here.

With labeled data, pairs with the same label are considered positive, and negative otherwise.
Here we consider unlabeled data sets.
One way to achieve this is to learn such pairs directly from Euclidean proximity, e.g., by "labeling" points $x_i, x_j$ positive if $\|x_i - x_j\|$ is small or taking the nearest neighbors of each point throughout the whole data set and negative otherwise.
Such a Siamese network can then be trained by minimizing the contrastive loss,
\begin{equation}
\begin{aligned}
L_0 &= \sum_{(\mathbf{x}_i, \mathbf{x}_j) \in \mathcal{P}^+}\left\|\mathbf{x}_i - \mathbf{x}_j\right\|_2^2 \\
&+ \sum_{(\mathbf{x}_i, \mathbf{x}_j) \in \mathcal{P}^-} \max(\lambda-\left\|\mathbf{x}_i - \mathbf{x}_j\right\|_2,0)^2,
\label{siamese_loss}
\end{aligned}
\end{equation} 
where $\lambda$ is a margin, and $\mathcal{P}^+, \mathcal{P}^-$ denotes positive and negative pair sets.

There exist basic problems for the above concept to be used in the scenario considered here:
(a) we cannot define positive or negative pairs from raw data, due to the variable length of the audio segments and the disturbance caused by of speaker characteristics, and
(b) it is time-consuming to find nearest neighbors for each point, since a large amount of data points is required for training audio embeddings.
These problems will be taken care of below.

\subsection{Phonetic Embedding with Speaker Characteristics Disentangled} \label{subsec:stage1}

\begin{figure}[t]
  \centering
  \includegraphics[width=\linewidth]{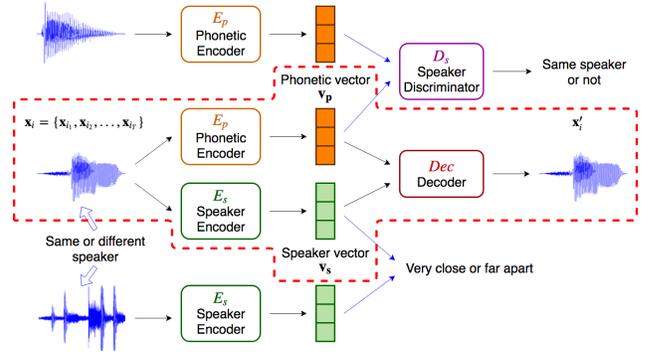}
  \caption{Phonetic embedding with speaker characteristics disentangled.}
  \label{fig:autoencoder}
\end{figure}

Audio embedding is to represent each audio segment for a linguistic unit (a word, syllable or phoneme) as a vector of a fixed dimensionality. 
This partly solves the first problem mentioned above, i.e., the audio segments have variable lengths.
Even with the fixed dimensionality, we note a linguistic unit with a given phonetic content corresponds to infinite number of audio realizations with varying acoustic factors such as speaker characteristics, microphone characteristics, background noise, etc.
All the latter acoustic factors are jointly referred to as speaker characteristics here for simplicity, which obviously disturb the goal of embedding signals for the same linguistic unit to vectors very close to each other.
This is why we wish to disentangle such factors here.

As shown in the middle of Figure~\ref{fig:autoencoder} following exactly the prior work~\cite{chen2018phonetic}, a sequence of acoustic features $\mathbf{x}_i=(\mathbf{x}_{i_1}, \mathbf{x}_{i_2}, ..., \mathbf{x}_{i_T})$ is entered to a phonetic encoder $E_p$ and a speaker encoder $E_s$ to obtain a phonetic vector $\mathbf{v_p}$ in orange and a speaker vector $\mathbf{v_s}$ in green. 
Then the phonetic and speaker vectors $\mathbf{v_p}$, $\mathbf{v_s}$ are used by the decoder $Dec$ together to reconstruct the acoustic features $\mathbf{x}'$. 
This phonetic vector $\mathbf{v_p}$ will be used as the phonetic embedding, or the audio embedding considered here carrying primarily the phonetic information in the signal.
The two encoders $E_p$, $E_s$ and the decoder $Dec$ are jointly learned by minimizing the reconstruction loss.

The training of the speaker encoder $E_s$ requires speaker information for the audio segments.
Assume the audio segment $\mathbf{x}_i$ is uttered by speaker $s_i$.
When the speaker information is not available, we can simply assume that the audio segments in the same utterance are produced by the same speaker. 
As shown in the lower part of Figure~\ref{fig:autoencoder}, $E_s$ is learned to minimize the contrastive loss. 
That is, if $\mathbf{x}_i$ and $\mathbf{x}_j$ are uttered by the same speaker ($s_i = s_j$), we want their speaker embeddings $\mathbf{v_s}_i$ and $\mathbf{v_s}_j$ to be as close as possible.
But if $s_i \neq s_j$, we want the distance between $\mathbf{v_s}_i$ and $\mathbf{v_s}_j$ larger than a threshold.

As shown in the upper right corner of Figure~\ref{fig:autoencoder}, a speaker discriminator $D_s$ takes two phonetic vectors $\mathbf{v_p}_i$ and $\mathbf{v_p}_j$ as input and tries to tell if the two vectors come from the same speaker.
The learning target of the phonetic encoder $E_p$ is to "fool" this speaker discriminator $D_s$, keeping it from discriminating the speaker identity correctly. 
In this way, only the phonetic information is learned in the phonetic vector $\mathbf{v_p}$, while only the speaker characteristics is encoded in the speaker vector $\mathbf{v_s}$.

\subsection{Identify Positive and Negative Pairs within each Mini-Batch}
\label{subsec:batch-pairs}

Finding nearest neighbors for each data point is costly, with time complexity of $\mathcal{O}(M^2)$, where $M$ is the corpus size.
To reduce the time and computing costs, we alternatively create a $k$-nearest neighbors graph among all the data points in each mini-batch, and use it to approximate the distribution for the full data set.
In this way, time complexity could be reduced to $\mathcal{O}(\|\mathcal{B}\|^2\cdot M / \|\mathcal{B}\|) = \mathcal{O}(M\|\mathcal{B}\|)$, where $\|\mathcal{B}\|$ is the mini-batch size.

\subsection{Siamese Style Training}
\label{subsec:siamese-style}

We can simply apply the Siamese style loss function as an extra requirement in training the phonetic embedding in subsection~\ref{subsec:stage1}, or the Siamese requirement is jointly trained:
\begin{equation}
\begin{aligned}
L_1 &= \sum_{(\mathbf{v_p}_i, \mathbf{v_p}_j) \in \mathcal{P_B}^+}\left\|\mathbf{v_p}_i - \mathbf{v_p}_j\right\|_2^2 \\
&+ \sum_{(\mathbf{v_p}_i, \mathbf{v_p}_j) \in \mathcal{P_B}^-} \max(\lambda-\left\|\mathbf{v_p}_i - \mathbf{v_p}_j\right\|_2,0)^2,
\label{siamese_loss_3}
\end{aligned}
\end{equation} 
where $\mathcal{P_B}^+, \mathcal{P_B}^-$ are positive and negative pairs selected in each mini-batch, and $\mathbf{v_p}_i, \mathbf{v_p}_j$ are the phonetic embeddings obtained in this way.

On the other hand, we can also pretrain the audio embeddings with speaker characteristics disentangled as in subsection 2.2, then on top of the obtained phonetic embeddings $\mathbf{v_p}$, train another Siamese model to further transform them to a new space where the similar points are more compact by clustered, or adjacency-based clustering.
The training loss function for this extra model is:
\begin{equation}
\begin{aligned}
L_2 &= \sum_{(\mathbf{v_p}_i, \mathbf{v_p}_j) \in \mathcal{P_B}^+}\left\|\mathbf{z}_i - \mathbf{z}_j\right\|_2^2 \\
&+ \sum_{(\mathbf{v_p}_i, \mathbf{v_p}_j) \in \mathcal{P_B}^-} \max(\lambda-\left\|\mathbf{z}_i - \mathbf{z}_j\right\|_2,0)^2,
\label{siamese_loss_2}
\end{aligned}
\end{equation} 
where the phonetic embeddings obtained in subsection~\ref{subsec:stage1}, $\mathbf{v_p}_i, \mathbf{v_p}_j$, are transformed to the new embeddings $\mathbf{z}_i, \mathbf{z}_j$.

\subsection{Dealing with Unbalanced Data}
\label{subsec:unbalanced-fix}

The distribution of linguistic units is unbalanced.
For low frequency units we may not be able to find more than two audio segments in a mini-batch. 
When we create a $k$-nearest neighbor graph within the mini-batch, audio segments for such units would be forced to reduce their distance to audio segments corresponding to different linguistic units.
On the other hand, for high frequency units with more than $k$ corresponding audio segments in a mini-batch, such audio segments would be separated into two or more clusters.

Therefore, instead of creating a $k$-nearest neighbor graph for a mini-batch, we alternatively create a fully-connected graph for each mini-batch, labeling pairs of data points with top-$k$ shortest distance in between as the positive pairs, and randomly select other $k$ pairs of data points as negative pairs.
In this way the probability that data points for positive pairs correspond to the same linguistic unit may be higher.

\section{Experimental Setup}
\label{sec:setup}

\subsection{Dataset}
\label{subsec:dataset}

We used LibriSpeech~\cite{panayotov2015librispeech} as the audio corpus in the experiments, which is a corpus of read speech in English derived from audiobooks. This corpus contained 1000 hours of speech sampled at 16 kHz uttered by 2484 speakers.
We randomly sampled 100 speakers from the ``clean" and ``others" sets, about 40 hours of speech for training, another 40 hours for testing, and 39-dim MFCCs were extracted as the acoustic features . 
The audio signals were segmented into three levels of linguistic units, word, syllable, and phoneme.

\subsection{Model Implementation}
\label{subsec:implementation}

The phonetic encoder $E_p$, speaker encoder $E_s$ and decoder $Dec$ were either 2-layer bi-directional GRUs or 3-layer CNNs with dense layers.
The size of embedding vectors is 256.
The speaker discriminator $D_s$ is a fully-connected feedforward network with 2 hidden layers with size 128.
The value of $\lambda$ we used in Eqs (\ref{siamese_loss_3}) (\ref{siamese_loss_2}) was set to 1.

\section{Experimental Results}
\label{sec:exp}

In the following subsections, we evaluate four kinds of audio embeddings: (a) Audio Word2Vec~\cite{chung2016audio}, which is simply an auto-encoder; (b) Audio Word2Vec with speaker characteristics disentangled as in subsection~\ref{subsec:stage1}~\cite{chen2018towards,chen2018phonetic}; (c) proposed approach as in Eq.(\ref{siamese_loss_3}); and (d) proposed approach as in Eq.(\ref{siamese_loss_2}).

\subsection{Analysis of Embedding Characteristics}
\label{subsec:exp-cluster}

We first compared the averaged cosine similarity of intra- and inter-class pairs for three different levels of linguistic units.
Intra-class pairs were evaluated between audio segments corresponding to the same linguistic units, while inter-class pairs were evaluated between segments belonging to different units.
Except for the four kinds of audio embeddings mentioned, we also provided two kinds of audio embedding as baselines: audio embedding obtained in subsection~\ref{subsec:stage1} with minimizing overall L1 loss as an extra requirement in training process ((b)+L1), and audio embedding obtained in subsection~\ref{subsec:stage1} with minimizing overall L2 loss as an extra requirement in training process ((b)+L2).

The results are listed in Table~\ref{table:avg-cos-sim}. 
It can be clearly found that row (d) for the proposed approach of Eq.(\ref{siamese_loss_2}) gave the highest intra-class average cosine similarity and the maximum difference between intra- and inter-class average cosine similarity for all the three linguistic units, which indicated that this approach offered better clustered audio embeddings.
In other words, those corresponding to the same linguistic units were more compactly distributed even without extra annotation.

\begin{table}[t]
\scriptsize
\centering
\caption{Average cosine similarity of intra- and inter-class pairs for three different levels of linguistic units.}
\label{table:avg-cos-sim}
\begin{tabular}{|c|cc|c|cc|c|cc|c|}
\hline

\multicolumn{1}{|c|}{\multirow{2}{*}{}} &
\multicolumn{3}{c|}{word} &
\multicolumn{3}{c|}{syllable} &
\multicolumn{3}{c|}{phoneme} \\ \cline{2-10}

& intra & inter & $\Delta$ & intra & inter & $\Delta$ & intra & inter & $\Delta$ \\ \hline \hline

(a) & .104 & .036 & .068 & .139 & .050 & .089 & .135 & \bf{-.003} & .138 \\ \hline
(b) & .083 & .024 & .059 & .131 & .039 & .092 & .088 & -.003 & .091 \\ \hline
(b)+L1  & .085 & \bf{.017} & .068 & .112 & .038 & .074 & .102 & -.003 & .105 \\ \hline
(b)+L2  & .078 & .024 & .054 & .107 & .037 & .070 & .090 & -.003 & .093 \\ \hline
(c) & .074 & .030 & .044 & .096 & \bf{.032} & .064 & .110 & .011 & .099 \\ \hline
(d) & \bf{.222} & .022 & \bf{.200} & \bf{.236} & .035 & \bf{.201} & \bf{.245} & .051 & \bf{.194} \\ \hline

\end{tabular}
\end{table}

\subsection{Analysis of Unsupervised Clustering}

\begin{figure}[t]
  \centering
  \includegraphics[width=\linewidth]{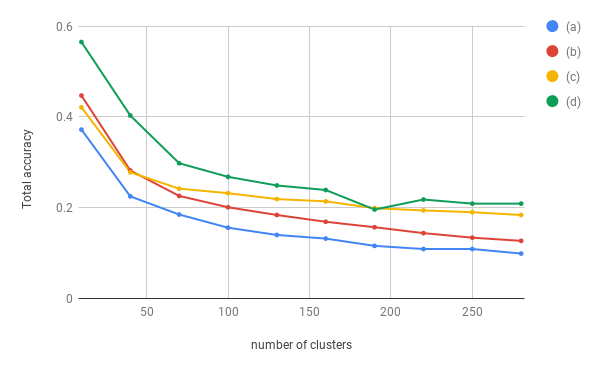}
  \caption{Total accuracy $Acc$ in Eq.(\ref{eq:max_acc}) for clustering characteristics analysis: (a) Audio Word2Vec, (b) Audio Word2Vec with speaker characteristics disentangled as in subsection~\ref{subsec:stage1}, (c) proposed with Eq.(\ref{siamese_loss_3}), (d) proposed with Eq.(\ref{siamese_loss_2}). Results for different levels of linguistic units have similar trends, so only results for of words are shown.}
  \label{fig:precision}
\end{figure}

In the experiment here, we use k-means, an unsupervised clustering method, for the first experiment for analysis. 
All three different levels of linguistic units were tested for comparison, so the numbers of labels were fixed to 70, which is the total class number of phoneme.
For word and syllable, the top 70 frequent units were selected as labels for experiments.

Given clustering results, we could construct a confusion matrix $C \in \mathcal{N}^{m \times n}$, where $m$ is the number of linguistic units tested (70), $n$ is the number of clusters (tested up to 280), and $C[i,j]$ indicates the count of data points having label $i$ but assigned to cluster $j$.
This count was first normalized,
\begin{equation}
\begin{aligned}
c(i, j) = \frac{C[i, j]}{\sum_i \sum_j C[i, j]}. 
\label{eq:acc}
\end{aligned}
\end{equation} 
For each label $i$ we obtained a cluster $j^*$ yielding the highest $c(i,j)$ and assumed it was the cluster for label $i$,
\begin{equation}
\begin{aligned}
j^* = \argmax_j c(i,j).
\label{eq:j_star}
\end{aligned}
\end{equation} 
The total accuracy was then evaluated by summing $c(i, j^*)$ over all labels $i$, 
\begin{equation}
\begin{aligned}
Acc = \sum_{i=1}^m c(i, j^*).
\label{eq:max_acc}
\end{aligned}
\end{equation} 
Higher total accuracy would be obtained if all the data points of the same label were in the same cluster.

The results of three different levels of linguistic units are similar, so only the clustering results of word are shown in Fig.~\ref{fig:precision} for clarity.
It is known that the clustering performance of k-means depends on both the number of clusters $n$, so various values of $n$ were tested for overall results.
First of all, we found that the feature disentanglement improved the total accuracy (curves (b) v.s. (a)), and the proposed Siamese style training in Eq.(\ref{siamese_loss_2}) gave further progress (curves (d) v.s. (a)).
As shown in Fig.~\ref{fig:precision}, curve (d) for the proposed approach of Eq.(\ref{siamese_loss_2}) gave the highest total accuracy at nearly all values of $n$, which proved that this approach greatly improved the performance in unsupervised clustering.

\subsection{Analysis of Spoken Term Detection}
\label{subsec:exp-retrieval}

\begin{table}[t]
\scriptsize
\centering
\caption{Spoken term detection with 80 queries.}
\label{table:retrieval}
\begin{tabular}{|c|c|c|c|c||c|c|}
\hline

\multicolumn{1}{|c|}{top\_$k$} &
\multicolumn{1}{c|}{(a)} &
\multicolumn{1}{c|}{(b)} &
\multicolumn{1}{c|}{(c)} &
\multicolumn{1}{c||}{(d)} &
\multicolumn{1}{c|}{(d) - (a)} &
\multicolumn{1}{c|}{(d) - (b)} \\ \hline \hline

1   & 33.59\% & 33.53\% & 33.68\% & \bf{33.88}\% & 0.29\% & 0.35\% \\ \hline
5   & 34.34\% & 34.39\% & 34.59\% & \bf{35.13}\% & 0.79\% & 0.74\% \\ \hline
10  & 34.70\% & 34.79\% & 35.05\% & \bf{35.67}\% & 0.97\% & 0.88\% \\ \hline
20  & 35.10\% & 35.23\% & 35.53\% & \bf{36.20}\% & 1.10\% & 0.97\% \\ \hline
40  & 35.52\% & 35.70\% & 36.05\% & \bf{36.71}\% & 1.19\% & 1.01\% \\ \hline
60  & 35.78\% & 36.00\% & 36.37\% & \bf{37.00}\% & 1.22\% & 1.00\% \\ \hline
                                             

\end{tabular}
\end{table}

We used the 960 hours of ``clean" and ``other" parts of LibriSpeech data set as the target archive for detection, which consisted of 1478 audio books with 5466 chapters. 
Each chapter included 1 to 204 utterances or 5 to 6529 spoken words.
In our experiments, 80 queries were chosen from the words used in these 960 hours of speech with the top 80 TF-IDF scores, and the chapters were taken as the spoken documents to be retrieved.
The audio realization of each query was randomly sampled from LibriSpeech data set, and our goal was to retrieve documents containing those queries (words, not necessarily the exact audio realizations). 
We used mean average precision (MAP) as the evaluation metric for the spoken term detection test.

For each query $q$ and each document $d$, the relevance score of $d$ with respect to $q$, $s(q,d)$, is defined as follows:
\begin{equation}
\begin{aligned}
s(q, d) &= \frac{1}{k}\sum_{w \in S_k(d, q)} cos(R(w), R(q)),
  \label{score}
\end{aligned}
\end{equation}
where $R(w)$ is the audio embedding of a spoken word $w$, $cos(\cdot)$ represents cosine similarity, and $S_k(d, q)$ is the set of top $k$ spoken words $w$ in $d$ with the highest cosine similarity values $cos(R(w), R(q))$, $k$ is a parameter. 
In other words, the documents $d$ were ranked by the average of top $k$ cosine similarity between each spoken word $w$ in $d$ and the query $q$.

The results are listed in Table~\ref{table:retrieval}.
As can be found from this table, colomn (d) for the proposed approach of Eq.(\ref{siamese_loss_2}) offered the best detection performance than all the other kinds of audio embedding at all values of $k$.
Because the queries are high frequency terms and they usually appear with multiple times in the documents, the detection performances of all kinds of audio embeddings gradually improved as $k$ increased in the range tested.
The right most two columns also listed the differences between the proposed approach in Eq.(\ref{siamese_loss_2}) (column (d)) and columns (a) and (b).
We see the proposed method gained large improvements.
As shown in the table, $k$ at 40 gave the maximum difference between columns (d) and (b).
This is probably related to the fact that the number of utterances in a document is roughly 40 in average.
Larger difference was achieved as $k$ was increased from 1 to 40, but less as $k$ was increased over 40.

\section{Conclusions and Future Work}
\label{sec:conclusion}

In this paper we propose a framework to embed audio segments into better clustered vector representations with fixed dimensionality, including identifying positive and negative pairs from unlabeled data for Siamese style training, disentangling acoustic factors such as speaker characteristics from the audio embedding, handling unbalanced data distribution.
Our proposed methods gave great improvement in both clustering analysis and spoken term detection.
For the future work, we have committed ourselves to distilling only linguistic information from audio segments.


\bibliographystyle{IEEEbib}
\bibliography{mybib,newbib,IR_bib,ref_dis,segment,transfer,INTERSPEECH16,ICASSP13}

\end{document}